\documentclass[final]{anthology-ch}         

\usepackage{booktabs}
\usepackage{float}
\usepackage{graphicx}
\usepackage{placeins}
\usepackage{siunitx}
\usepackage{tabularx}
\usepackage{threeparttable}
\usepackage{wrapfig}

\title{A Free Lunch? Adapting PP-OCRv6 for Historical Text Recognition}

\author[1]{Benjamin Kiessling}[
  orcid=0000-0001-9543-7827
]

\affiliation{1}{ALMAnaCH, Inria Paris}

\keywords[eng]{automatic text recognition, historical documents, handwritten text recognition, optical character recognition, PP-OCRv6}

\pubyear{2025}
\pubvolume{1}
\pagestart{1}
\pageend{1}
\paperorder{1}
\conferencename{Proceedings of Conference XXX}
\conferenceeditors{Editor1 Editor2}
\doi{00000/00000}  
\customhead{}  

\begin{document}

\maketitle

\begin{abstract}
Despite impressive reported scores, large vision-language models have seen
limited practical uptake in historical automatic text recognition because of
their computational cost, dependence on large-scale pretraining, and
hallucination. Historical ATR therefore continues to rely largely on compact
CRNN line recognizers, which are visually grounded and trainable on modest
data. Lightweight recurrence-free recognizers promise the accuracy of larger
models with the practical advantages of CRNNs, yet have not been
comprehensively evaluated on historical writing. We adapt PP-OCRv6, a recent
compact text recognizer without strong language modeling, for historical line
recognition and compare it with a conventional CRNN across generalized
pretraining, domain-specific training, corpus-level fine-tuning, and
manuscript-specific few-shot adaptation on multilingual Latin- and
Arabic-script material. While PP-OCRv6 does not consistently outperform the
baseline when trained from scratch, heterogeneous pretraining produces
markedly better generalization. Comparisons with the Qwen3.5-based Medusa
recognizer further show that fine-tuned PP-OCRv6 can outperform a large VLM
tailored towards historical Latin-script HTR.
\end{abstract}

\section{Introduction}
\label{sec:introduction}

For the better part of a decade, line-wise recognition with convolutional
recurrent neural networks (CRNNs) trained using connectionist temporal
classification (CTC)~\cite{graves2006ctc} has been the dominant paradigm in
automatic text recognition (ATR), including for historical material.
Typically combining a convolutional feature extractor, bidirectional long
short-term memory layers, and a linear projection
\cite{shi2016crnn,puigcerver2017multidimensional}, these systems are accurate,
robust, trainable on the modest amounts of annotated data available for
individual historical corpora, and require no character-level segmentation.
For these reasons, they underlie widely used recognition engines such as
Calamari~\cite{wick2018calamari}, Tesseract
4~\cite{tesseract2018neural}, and kraken~\cite{kiessling2025kraken}.

Text recognition has recently shifted towards large vision-language models.
Besides achieving excellent error rates in conventional line recognition,
these models can recognize entire paragraphs or pages without explicit
segmentation. Unlike earlier methods such as Start, Follow,
Read~\cite{tensmeyer2019sfr}, which learn line extraction and recognition
jointly, Qwen-VL~\cite{bai2023qwenvl} and
LightOnOCR~\cite{taghadouini2026lightonocr} directly produce ordered text and
document structure from complete page images. Dispensing with line segmentation
is especially attractive for historical ATR, where segmentation errors are
frequently responsible for a significant proportion of overall recognition
errors.

Generative recognition, in particular in the full-page setting, has
nevertheless seen limited adoption in practical historical ATR. Autoregressive
systems with strong language models can produce insufficiently grounded
transcriptions, deriving text from linguistic priors without corresponding
visual evidence. This and similar phenomena, commonly called hallucination, are
still poorly understood. Recent work on Ancient Greek editions has demonstrated
extensive overcorrection even in the presence of extensive changes such as word
shuffling~\cite{karamolegkou2026reading}, pointing towards a fundamental
problem in the construction of these systems.

Large autoregressive models are also expensive to train and deploy, and acquire
their capabilities from pretraining corpora that cannot reasonably be
produced for most historical collections. The mismatch is especially acute
for non-Western traditions, obsolete scripts, minority and low-resource
languages, and projects using specialized transcription norms. Synthetic data
can alleviate annotation scarcity, for example by rendering transcriptions in
digital typefaces approximating the target hands~\cite{barrere2024fewdata}.
This approach remains limited to writing with suitable typographic surrogates,
which places substantial portions of the written record beyond the practical
reach of models that rely on large-scale pretraining.

These drawbacks have renewed interest in compact recognizers that replace
recurrent layers without introducing high-capacity generative language models.
Such systems retain the modest data requirements and visual grounding of
conventional line recognition while permitting parallel execution on modern
hardware. PP-OCRv6 is a recent general-purpose example whose authors report
strong results on printed and handwritten text in document scans and natural
scenes, with models ranging from 1.1 to 19 million
parameters~\cite{zhang2026ppocrv6}.

This article asks whether PP-OCRv6 offers a ``free lunch'' for historical ATR
as a replacement for recurrent line recognition. The method is adapted for
historical line images and evaluated when trained from scratch on
project-specific data, pretrained as a generalized model, and fine-tuned
towards particular datasets and transcription norms. Comparisons with kraken
and Medusa, a Qwen3.5-9B-based line recognizer
\cite{moins2026medusa}, examine recognition accuracy, few-shot adaptation, and
inference throughput.

\section{Related Work}
\label{sec:related-work}

\subsection{Recurrent line recognition with CTC}
\label{sec:related-recurrent}

Connectionist temporal classification permits alignment-free sequence modelling
and therefore permits training from line images and their transcriptions
without character-level annotations~\cite{graves2006ctc}. Combined with a
convolutional feature extractor and bidirectional LSTMs, it has defined
historical machine-printed and handwritten text recognition for the last
decade; representative examples include the systems described by
\cite{puigcerver2017multidimensional} and \cite{wick2018calamari}. Although
highly accurate, their recurrent sequence processing cannot fully exploit the
hardware and software optimizations available to convolutional and attentional
models.

\subsection{Language-generative recognition}
\label{sec:related-generative}

The principal attraction of language-generative recognition is its promise of
higher recognition accuracy through the combination of visual features and
strong pretrained language models. Representative systems include TrOCR,
which combines pretrained image and text
Transformers~\cite{li2023trocr}, and DTrOCR, whose decoder is initialized from
GPT-2~\cite{fujitake2023dtrocr}; both operate on presegmented text images.

Generative models also offer greater flexibility with respect to segmentation.
Removing line segmentation as a fragile component of the recognition pipeline
has been a longstanding objective. Start, Follow, Read learns line localization
and recognition jointly~\cite{tensmeyer2019sfr}, OrigamiNet unfolds the
two-dimensional page into a sequence~\cite{yousef2020origaminet}, and DAN uses
attention to traverse the document~\cite{coquenet2022dan}. These pre-VLM
systems reconstruct an internal segmentation through task-specific
architectures that are brittle and difficult to train. Large VLMs instead use
a common autoregressive formulation across text lines, paragraphs, and
complete pages. Qwen-VL provides this flexibility as part of a general-purpose
prompted model~\cite{bai2023qwenvl}; LightOnOCR applies it specifically to
complete-page recognition~\cite{taghadouini2026lightonocr}; and Churro is
designed specifically for full-page historical
ATR~\cite{semnani2025churro}.

\subsection{Compact recurrence-free recognizers}
\label{sec:related-compact}

Attempts to eliminate recurrence long predate the widespread adoption of large
Transformer models. \cite{such2018fully} explored fully convolutional sequence
decoding, while \cite{gao2019reading} and \cite{yousef2020accurate} retained
CTC but replaced recurrent sequence modelling with convolutions. Despite strong
in-domain results, these models often generalized less well than recurrent
systems in practical transfer. This remained obscured by limited in-domain
evaluation, which is a poor predictor of performance under domain
shift~\cite{garrido2025generalization}.

More recent recurrence-free recognizers include the convolutional Gated Fully
Convolutional Network~\cite{coquenet2020gfcn} and
Easter2.0~\cite{chaudhary2022easter}. Attentional designs include the Modular
Light Transformer~\cite{barrere2022light}, HTR-VT~\cite{li2025htrvt}, and
SVTRv2~\cite{du2025svtrv2}. Their scale remains far below that of
language-generative recognizers: the Modular Light Transformer has 7.7 million
parameters~\cite{barrere2022light}, HTR-VT 53.5 million~\cite{li2025htrvt}, and
PP-OCRv6 1.1-19 million, compared with one billion for LightOnOCR, itself
considered a lightweight VLM.

\section{Methodology}
\label{sec:methodology}

PP-OCRv6 is a two-stage OCR system for document scans and natural scene
images. As only its potential for replacing a CRNN baseline is evaluated, only
its recognition stage is considered.

\subsection{PP-OCRv6 text recognition}
\label{sec:method-ppocr}

The recognition component combines an LCNetV4 convolutional backbone, an
attentional LightSVTR neck, and a CTC head. The tiny variant omits LightSVTR,
while the small and medium variants increase backbone capacity and neck width.
The generalized recognizers reported by the authors support Chinese and contain
1.1, 5.2, and 19 million parameters. The smaller character inventories used
in our experiments reduce the size of the final linear projections and the respective
parameter counts to 0.68, 3.22, and 15.86 million.

Training combines equally weighted CTC and label-smoothed cross-entropy from an
auxiliary autoregressive NRTR decoder~\cite{sheng2019nrtr}. The decoder acts as
a regularizer and is discarded for inference, which remains CTC-based. The
original method uses the Adam optimizer and a canonical input size of
$48\mathbin{\times}320$ pixels.

PP-OCRv6 inherits a data curation procedure from PP-OCRv5~\cite{cui2026ppocrv5}
intended to filter noisy, incorrect, and trivially easy transcriptions. A
referee recognizer assigns each line in the training dataset a mean character
confidence. Samples below 0.80 are treated as difficult or noisy and those
above 0.97 as trivial; around half of the original training set falls within
the 0.95-0.97 confidence range the authors determine to be optimal.

\subsection{Adaptation to historical text recognition}
\label{sec:method-adaptation}

The nominal PP-OCRv6 input line height is doubled from 48 to 96 pixels to
preserve fine strokes, diacritics, and abbreviation marks in degraded
historical text. In preliminary experiments, further increases did not result
in additional validation score gains. This intervention is consistent with
other systems optimized for historical ATR: 128 pixels in
PyLaia~\cite{pylaia2024format} and 120 in kraken~\cite{kiessling2025kraken},
compared with 36 pixels in Tesseract 4~\cite{tesseract2018training}, which is
optimized for modern machine-printed text.

The default PP-OCRv6 configuration imposes two restrictive limits that are
manifestly unsuitable for historical ATR: NRTR targets are capped at 23 code
points, and every line image is resized to $48 \times 320$ pixels. At the
model's horizontal stride of eight, the latter provides only 40 CTC steps,
which is insufficient for historical lines of even moderate length. Lines are
instead scaled to 96 pixels in height while preserving their aspect ratio, so
that the available CTC sequence length grows with the input line. The NRTR
target limit is raised accordingly.

The distillation loss using a medium teacher model for the original PP-OCRv6
tiny variant is omitted.

Confidence-window filtering is also discarded. Its lower boundary was tested
on the French, Latin, and Spanish subset of the generalized-model training
corpus, where it rejects 17\% of lines, compared with approximately 3\% of the
original training data~\cite{cui2026ppocrv5}. The medium-variant generalized
model obtains a CER of at most 15\% on 52\% of the rejected material. The
correctly annotated examples in Figure~\ref{fig:confidence-filtering} receive
filter confidences between 0.55 and 0.68 despite being reproduced exactly by
the generalized model.

Confidence alone is therefore insufficient for detecting incorrect annotations
because domain familiarity and annotation quality are intermingled in a single
score. On heterogeneous historical data, confidence filtering removes difficult
but correctly annotated lines and underrepresents rare scripts, hands, and
typographies. It is therefore not used for pretraining.

\providecommand{\htrfont}{}
\providecommand{\confrejectline}[3]{%
  \noindent\includegraphics[width=\linewidth]{#1}%
  \par\nobreak\vskip1pt
  {\scriptsize\htrfont\raggedright #2\hfill $c=#3$\par}%
  \vskip3pt}
\begin{wrapfigure}{r}{0.42\textwidth}
  \centering
  \confrejectline{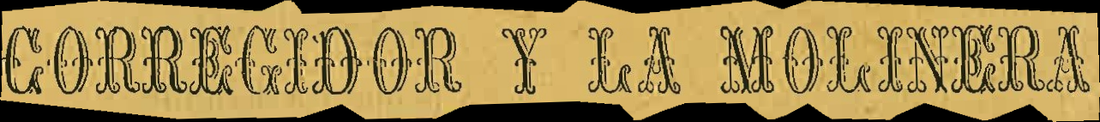}{CORREGIDOR Y LA MOLINERA}{0.55}
  \confrejectline{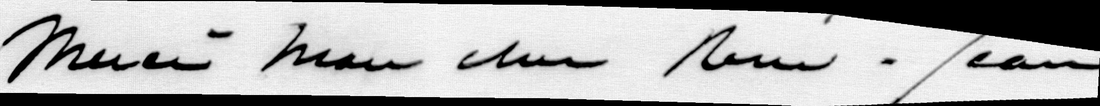}{Merci mon cher René-Jean}{0.59}
  \confrejectline{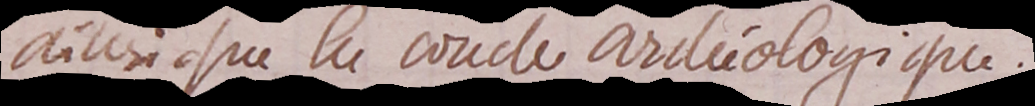}{ainsi que la couche archéologique.}{0.68}
  \confrejectline{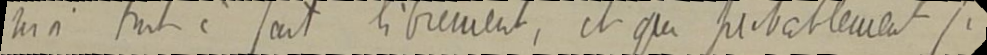}{moi tout à fait librement, et que probablement je}{0.68}
  \confrejectline{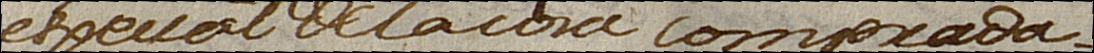}{especial de la cosa comprada.}{0.67}
  \confrejectline{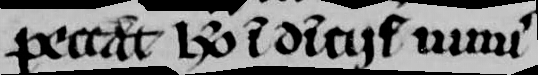}{peccat homo in diuiciis nimis}{0.62}
  \vskip-3pt
  \caption{Correctly annotated lines rejected by confidence-window filtering.}
  \label{fig:confidence-filtering}
\end{wrapfigure}

The Adam optimizer was replaced with a hybrid optimizer, applying
Muon~\cite{jordan2024muon} to hidden linear weights and AdamW to all remaining
parameters, initially with the aim of accelerating convergence.

\subsection{Experiments}

The adapted tiny, small, and medium PP-OCRv6 variants are evaluated under four
scenarios: \textbf{generalized-model training} for general-purpose,
off-the-shelf recognition; \textbf{training from scratch} on a domain-specific
corpus when no suitable base model is available; \textbf{fine-tuning} a
generalized model for the best project-specific accuracy; and \textbf{few-shot
adaptation} for the highest possible accuracy on a single manuscript.

They are compared with the default CRNN recognizer of the kraken ATR
software~\cite{kiessling2025kraken}, a four-layer convolutional network
followed by three 200-unit bidirectional LSTM layers trained with CTC. 
Training was performed on NVIDIA A40 GPUs, using two devices for the medium
from-scratch runs and one otherwise.

The generalized models were pretrained on a heterogeneous corpus combining
handwritten, typed, and machine-printed in 37 different languages (ca. 6.2
million lines), and synthetic material (ca. one million lines).\footnote{Model
weights and a full corpus inventory are available on Zenodo
\url{https://doi.org/10.5281/zenodo.21788403};
\url{https://doi.org/10.5281/zenodo.21788405}; and
\url{https://doi.org/10.5281/zenodo.21788410}.} Its held-out test split contains
approximately 83000 lines in 33 languages. Almost all CATMuS and MAKHZAN
training material is included in the pretraining corpus, while their evaluation
sets remain held out.

The principal domain-specific evaluation uses CATMuS
Medieval~\cite{clerice2024catmus}. After removal of 7000 lines for
validation, approximately 142000 lines from 300 manuscripts remain.
CATMuS is relatively large for a historical ATR dataset and uniform
domain-specific datasets are frequently substantially smaller. The training
lines are therefore divided into nested 25, 50, 75, 90, and 100\% tiers, each
retaining all manuscripts, emulating these smaller training corpora.

For from-scratch training, each model is trained for a number of steps
equivalent to 20 epochs on the complete corpus; the number of epochs is
increased inversely with tier size so that all runs process the same number of
line images. Fine-tuning is performed for eight epochs. Evaluation uses the
ICDAR 2026 Competition on Multilingual Medieval Handwriting Recognition
(CMMHWR) test set~\cite{kiessling:hal-05608045}: French, Latin, and Spanish
test transfer to languages represented in CATMuS (task~1), Occitan tests
transfer within the Romance family (task~2), and Czech tests cross-family
transfer (task~3). A randomly sampled test set used for quality estimation
during the creation of the CoMMA corpus~\cite{clerice2025comma}\footnote{The
test set is available on Zenodo at
\url{https://zenodo.org/records/anonymized}.} provides an additional, more
distant evaluation on medieval Latin and French material.

\begin{wraptable}{r}{0.51\textwidth}
\centering
\scriptsize
\setlength{\tabcolsep}{2pt}
\caption{Generalized-model performance.}
\label{tab:base-models}
\begin{tabularx}{\linewidth}{@{}Xrrrrr@{}}
\toprule
& & \multicolumn{2}{c}{CER} & \multicolumn{2}{c}{WER} \\
\cmidrule(lr){3-4}\cmidrule(lr){5-6}
Model & Params. (M) & Micro & Macro & Micro & Macro \\
\midrule
kraken CRNN &  4.64 & 10.84 & 12.41 & 34.63 & 39.23 \\
PP-OCRv6 tiny   &  0.68 & 14.25 & 15.24 & 43.52 & 47.35 \\
PP-OCRv6 small  &  3.22 &  9.33 &  9.80 & 30.38 & 33.91 \\
PP-OCRv6 medium & 15.86 &  \textbf{7.03} &  \textbf{7.15} & \textbf{23.36} & \textbf{26.24} \\
\bottomrule
\end{tabularx}
\end{wraptable}

Accuracy is also compared with Medusa, a Qwen3.5-9B line recognizer submitted
at CMMHWR~\cite{moins2026medusa}, as an example of the recognition capabilities
of current large VLMs. Medusa is evaluated through the authors' DocWorkflow
pipeline, including its CATMuS-specific post-processing. VLMs often struggle
with historical hands and diplomatic transcription norms in zero-shot settings,
therefore a system that has been tuned towards the evaluated corpora provides a
fairer comparison.

To test whether the observed transfer behaviour is specific to Latin script,
e.g., due to a larger and more diverse sample of Latin-script material in the
pretraining dataset, zero-shot and fine-tuning evaluation is repeated on the
Arabic, Persian, and Ottoman Turkish subsets of OpenITI
MAKHZAN~\cite{allen2026makhzan}, a corpus of historical handwritten
Arabic-script material. In contrast to the medieval Latin-script evaluation,
the test set is not document-disjoint due to the construction of the
pretraining corpus.

\begin{wraptable}{R}{0.51\textwidth}
\centering
\scriptsize
\setlength{\tabcolsep}{3pt}
\caption{Architecture adaptation ablation (task~1 macro CER, \%).}
\label{tab:adaptation-ablation}
\begin{tabularx}{\linewidth}{@{}Xrrr@{}}
\toprule
Variant & 48 px & 96 px & 96 px + AdamW/Muon \\
\midrule
PP-OCRv6 tiny   & 24.05 & 12.57 & 13.13 \\
PP-OCRv6 small  & 19.28 & 10.54 & 10.50 \\
PP-OCRv6 medium & \textbf{18.35} & \textbf{10.10} &  \textbf{9.77} \\
\bottomrule
\end{tabularx}
\end{wraptable}

Few-shot manuscript-specific adaptation was evaluated separately on four
CMMHWR test pages using between 20 and 240 training lines. Each page was
divided into a training pool and a fixed same-page evaluation set. Up to five
disjoint samples were drawn at each budget, depending on the size of the
training pool, and each sample was used in five training runs. Every run
received 300 optimizer steps.

In all experiments, CER and WER are computed after NFD normalization and the
normalization of all whitespace to \texttt{U+0020}. Runtime is measured on
identical line images using an NVIDIA A40 and an Intel Xeon Silver 4316.

\section{Analysis}
\label{sec:analysis}

\subsection{Generalized models}

The generalized models are first evaluated on the held-out 33-language
pretraining test set. As shown in Table~\ref{tab:base-models},
PP-OCRv6 small and medium outperform the kraken baseline in both character and
word error rate, whereas the tiny variant performs worse. The progression of
error rates from tiny to medium is substantial, reducing macro CER from 15.2\%
to 7.2\%, which indicates that model capacity is the primary limitation when
learning from a highly heterogeneous corpus.

\subsection{CATMuS, CMMHWR, and CoMMA}

The effects of the adaptations to PP-OCRv6 explained above are shown in
Table~\ref{tab:adaptation-ablation}. As removal of confidence-window filtering
and fixed-width resizing is a prerequisite for historical ATR, only the line
height and optimizer change were ablated.

Increasing the line height accounts for almost all of the improvement, reducing
CMMHWR task~1 macro CER by 8.3-11.5 percentage points. Muon with AdamW was
initially adopted to accelerate convergence, but produces scores broadly in
line with pure Adam training. Adam training at 96 pixels was also found to be
less stable, whereas no corresponding failures were observed with the hybrid
optimizer. The 96-pixel configuration with Muon and AdamW is therefore used for
the remaining PP-OCRv6 experiments.

\begin{wraptable}{r}{0.64\textwidth}
\centering
\scriptsize
\setlength{\tabcolsep}{2pt}
\caption{CATMuS from-scratch results (CER; WER in parentheses, \%).}
\label{tab:catmus-scratch}
\begin{tabularx}{\linewidth}{@{}Xrrrr@{}}
\toprule
Test set & kraken CRNN & \multicolumn{3}{c}{PP-OCRv6} \\
\cmidrule(lr){3-5}
& & Tiny & Small & Medium \\
\midrule
\multicolumn{5}{@{}l}{\textit{CMMHWR}} \\
Task 1: French  & \textbf{4.32} {\scriptsize(\textbf{18.52})} & 7.34 {\scriptsize(31.23)} & 5.39 {\scriptsize(23.94)} & 4.94 {\scriptsize(22.22)} \\
Task 1: Latin   & \textbf{11.00} {\scriptsize(\textbf{44.92})} & 16.41 {\scriptsize(62.00)} & 13.32 {\scriptsize(54.06)} & 12.17 {\scriptsize(50.98)} \\
Task 1: Spanish & \textbf{11.85} {\scriptsize(\textbf{39.24})} & 15.65 {\scriptsize(49.35)} & 12.79 {\scriptsize(42.57)} & 12.21 {\scriptsize(42.17)} \\
Task 2: Occitan & \textbf{8.69} {\scriptsize(\textbf{39.55})} & 12.04 {\scriptsize(51.87)} & 9.70 {\scriptsize(44.82)} & 9.30 {\scriptsize(44.31)} \\
Task 3: Czech   & 26.67 {\scriptsize(\textbf{79.76})} & 28.16 {\scriptsize(81.61)} & \textbf{25.93} {\scriptsize(80.56)} & 25.99 {\scriptsize(80.33)} \\
\addlinespace
\multicolumn{5}{@{}l}{\textit{CoMMA}} \\
Latin  & \textbf{12.77} {\scriptsize(\textbf{45.98})} & 17.23 {\scriptsize(59.79)} & 14.07 {\scriptsize(52.49)} & 13.54 {\scriptsize(51.39)} \\
French & \textbf{22.63} {\scriptsize(\textbf{55.79})} & 27.62 {\scriptsize(65.40)} & 24.05 {\scriptsize(58.98)} & 24.31 {\scriptsize(58.59)} \\
\bottomrule
\end{tabularx}
\end{wraptable}

The full-corpus from-scratch results are reported in
Table~\ref{tab:catmus-scratch}. The kraken CRNN obtains lower CER and WER on
task~1, task~2, and CoMMA. PP-OCRv6 small and medium have marginally lower CER
on Czech, but not lower WER.

\begin{figure}[htbp]
\centering
\includegraphics[width=0.95\textwidth]{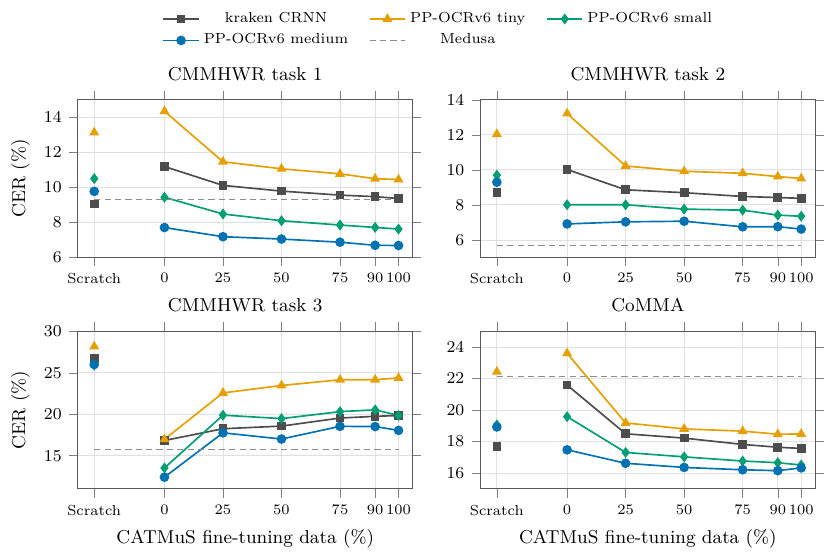}
\caption{CATMuS training and fine-tuning by evaluation set.}
\label{fig:catmus-adaptation}
\end{figure}

The effect of pretraining the generalized models and subsequent CATMuS fine-tuning is shown
in Figure~\ref{fig:catmus-adaptation}. The isolated markers give the scores of
models trained from scratch on the complete corpus; the connected series begin
with the zero-shot generalized models and continue over the five fine-tuning
tiers; the dashed line gives the Medusa result. Fine-tuning on only 25\% of
CATMuS improves every PP-OCRv6 variant over its counterpart trained from scratch
on the complete corpus across all seven test sets. More fine-tuning data
generally reduces error on task~1, task~2, and CoMMA. On Czech, however, every
generalized model performs best zero-shot and performance tends to deteriorate as
more CATMuS data are introduced. PP-OCRv6 medium also outperforms Medusa at this
point, with 12.4\% CER against 15.7\%, before CATMuS fine-tuning reverses the
ordering. This indicates forgetting of cross-family knowledge acquired during
pretraining.

Table~\ref{tab:catmus-adapted} compares the fully fine-tuned generalized models
with Medusa. PP-OCRv6 medium obtains lower CER and WER on task~1 and CoMMA,
while Medusa remains better on Occitan and Czech. The generalized-model
pretraining corpus contains no Occitan and only minuscule amounts of Czech,
while CATMuS contains neither language. Medusa, in contrast, was trained on
extensive Occitan and Czech material~\cite{moins2026medusa}, explaining its
better scores on tasks~2 and~3.

\begin{table}[htbp]
\centering
\setlength{\tabcolsep}{4pt}
\begin{threeparttable}
\caption{CATMuS-adapted models and Medusa (CER; WER in parentheses, \%).}
\label{tab:catmus-adapted}
\begin{tabularx}{\textwidth}{@{}Xrrrrr@{}}
\toprule
Test set & kraken CRNN & \multicolumn{3}{c}{PP-OCRv6} & Medusa\tnote{a} \\
\cmidrule(lr){3-5}
& & Tiny & Small & Medium & \\
\midrule
\multicolumn{6}{@{}l}{\textit{CMMHWR}} \\
Task 1: French  & 5.01 {\scriptsize(21.29)} & 6.26 {\scriptsize(27.52)} & 4.07 {\scriptsize(18.69)} & \textbf{3.36} {\scriptsize(\textbf{15.81})} & 3.74 {\scriptsize(15.91)} \\
Task 1: Latin   & 12.61 {\scriptsize(51.03)} & 13.19 {\scriptsize(54.22)} & 10.05 {\scriptsize(44.26)} & \textbf{8.70} {\scriptsize(\textbf{39.67})} & 13.07 {\scriptsize(43.96)} \\
Task 1: Spanish & 10.47 {\scriptsize(38.03)} & 11.91 {\scriptsize(41.50)} & 8.74 {\scriptsize(33.90)} & \textbf{7.99} {\scriptsize(\textbf{31.02})} & 11.19 {\scriptsize(32.74)} \\
Task 2: Occitan & 8.38 {\scriptsize(39.23)} & 9.51 {\scriptsize(44.42)} & 7.35 {\scriptsize(38.57)} & 6.62 {\scriptsize(36.03)} & \textbf{5.70} {\scriptsize(\textbf{21.91})} \\
Task 3: Czech   & 19.84 {\scriptsize(72.15)} & 24.37 {\scriptsize(78.92)} & 19.86 {\scriptsize(72.20)} & 18.05 {\scriptsize(70.29)} & \textbf{15.73} {\scriptsize(\textbf{64.94})} \\
\addlinespace
\multicolumn{6}{@{}l}{\textit{CoMMA}} \\
Latin  & 13.48 {\scriptsize(50.89)} & 13.83 {\scriptsize(53.19)} & 11.52 {\scriptsize(45.36)} & \textbf{11.30} {\scriptsize(\textbf{42.60})} & 16.08 {\scriptsize(45.54)} \\
French & 21.65 {\scriptsize(54.44)} & 23.14 {\scriptsize(58.66)} & 21.50 {\scriptsize(53.50)} & \textbf{21.36} {\scriptsize(\textbf{51.39})} & 28.24 {\scriptsize(55.74)} \\
\bottomrule
\end{tabularx}
\begin{tablenotes}[flushleft]
\footnotesize
\item[a] The task~1-3 CERs of 8.03\%, 5.24\%, and 10.8\% reported for
Medusa~\cite{moins2026medusa} could not be reproduced with the released
DocWorkflow pipeline. The values shown here were obtained with that pipeline
and its documented configuration.
\end{tablenotes}
\end{threeparttable}
\end{table}

\FloatBarrier
\subsection{Few-shot adaptation}

\begin{figure}[H]
\centering
\includegraphics[width=0.82\textwidth]{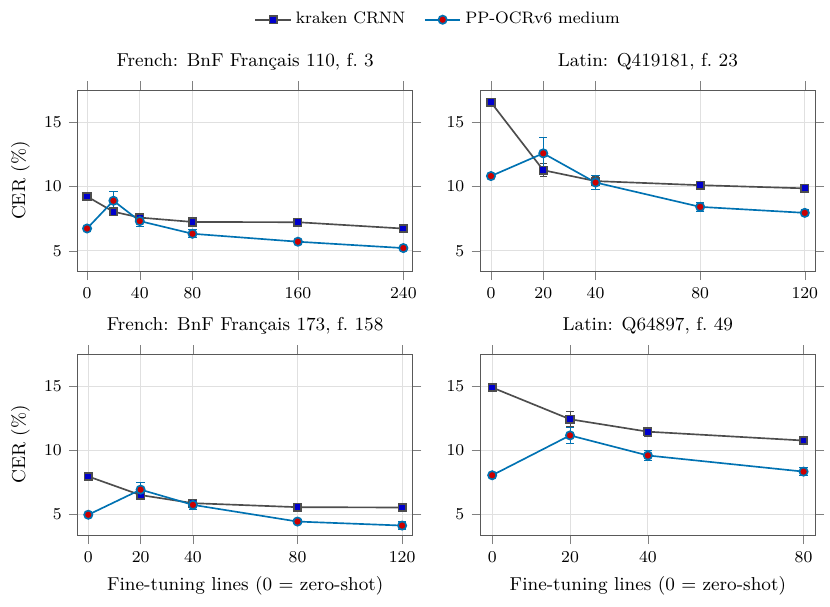}
\caption{Few-shot CER by manuscript and training-set size.}
\label{fig:fewshot}
\end{figure}

The few-shot results are shown in Figure~\ref{fig:fewshot}. A budget of zero
indicates zero-shot accuracy of the generalized model. Relative to this starting
point, the kraken baseline improves with only 20 training lines on all four
manuscripts. PP-OCRv6 instead performs worse after 20 lines, improves after 40
lines on one manuscript and after 80 lines on two others, and does not improve
within the maximum available budget of 80 lines on the fourth. This provides
limited evidence that the recurrent baseline is more effective in the extreme
few-shot regime. The slower improvement of PP-OCRv6 is offset by its lower
absolute CER, which is below that of the baseline on all four manuscripts from
40 training lines onwards.

Variation across data draws and training runs is small: with one exception, the
combined standard deviation remains below 0.8 percentage points and falls
below 0.4 percentage points from 80 training lines onwards. PP-OCRv6 is
nevertheless somewhat more variable than the baseline at the smallest budgets,
where it does not improve upon the base model.

\subsection{Arabic-script transfer}

Table~\ref{tab:makhzan} reports the MAKHZAN results by language and as an
unweighted macro average, reproducing the model ranking observed on the
Latin-script evaluations. PP-OCRv6 small and medium outperform the kraken
baseline in macro CER and WER both zero-shot and after fine-tuning, while the
tiny model does not.
PP-OCRv6 medium is already the strongest model zero-shot, and its macro CER
increases slightly from 18.0\% to 18.8\% after fine-tuning. The persistence
of this performance gap on Arabic-script material makes it unlikely that the
Latin-script results arise merely from an interaction between the larger
volume of Latin-script pretraining data and the greater effective capacity of
PP-OCRv6.

\begin{table}[H]
\centering
\setlength{\tabcolsep}{5pt}
\caption{MAKHZAN recognition results (CER; WER in parentheses, \%).}
\label{tab:makhzan}
\begin{tabularx}{\textwidth}{@{}Xrrrr@{}}
\toprule
Evaluation & kraken CRNN & \multicolumn{3}{c}{PP-OCRv6} \\
\cmidrule(lr){3-5}
& & Tiny & Small & Medium \\
\midrule
\multicolumn{5}{@{}l}{\textit{Generalized models, zero-shot}} \\
Arabic          & 24.46 {\scriptsize(70.97)} & 28.16 {\scriptsize(81.57)} & 20.67 {\scriptsize(66.02)} & \textbf{16.05} {\scriptsize(\textbf{52.95})} \\
Persian         & 36.42 {\scriptsize(86.84)} & 39.19 {\scriptsize(90.08)} & 30.35 {\scriptsize(80.53)} & \textbf{24.17} {\scriptsize(\textbf{70.32})} \\
Ottoman Turkish & 20.21 {\scriptsize(67.88)} & 22.57 {\scriptsize(73.72)} & 17.84 {\scriptsize(64.72)} & \textbf{13.76} {\scriptsize(\textbf{53.91})} \\
\addlinespace
Macro           & 27.03 {\scriptsize(75.23)} & 29.98 {\scriptsize(81.79)} & 22.95 {\scriptsize(70.43)} & \textbf{17.99} {\scriptsize(\textbf{59.06})} \\
\midrule
\multicolumn{5}{@{}l}{\textit{Fine-tuned on MAKHZAN}} \\
Arabic          & \textbf{22.23} {\scriptsize(66.61)} & 28.66 {\scriptsize(76.37)} & 25.37 {\scriptsize(70.80)} & 23.63 {\scriptsize(\textbf{65.94})} \\
Persian         & 34.20 {\scriptsize(84.32)} & 33.60 {\scriptsize(84.15)} & 27.71 {\scriptsize(75.88)} & \textbf{23.01} {\scriptsize(\textbf{67.52})} \\
Ottoman Turkish & 17.15 {\scriptsize(61.88)} & 15.74 {\scriptsize(58.73)} & 11.86 {\scriptsize(48.61)} & \textbf{9.83} {\scriptsize(\textbf{41.65})} \\
\addlinespace
Macro           & 24.52 {\scriptsize(70.94)} & 26.00 {\scriptsize(73.08)} & 21.65 {\scriptsize(65.09)} & \textbf{18.82} {\scriptsize(\textbf{58.37})} \\
\bottomrule
\end{tabularx}
\end{table}

\subsection{Inference speed}

\begin{figure}[H]
\centering
\includegraphics[width=0.78\textwidth]{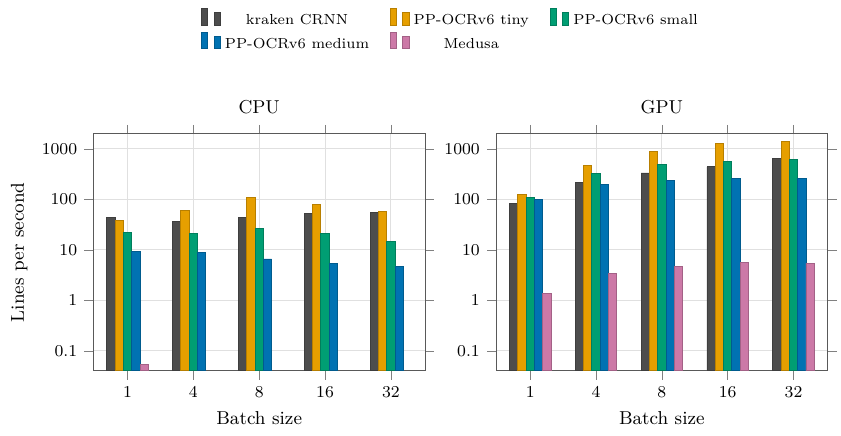}
\caption{Inference throughput by batch size.}
\label{fig:speed}
\end{figure}

With PP-OCRv6 small, the improvement in recognition accuracy does not incur a
substantial reduction in throughput on GPU. Its generalized model reduces macro
CER from 12.4 to 9.8\% and it outperforms the kraken CRNN on almost all test
sets after CATMuS fine-tuning. At the same time, it is faster up to batch size
16 and remains close at batch 32, as shown in Figure~\ref{fig:speed}, processing
approximately 610 instead of 640 lines per second.

The results for tiny and medium reflect their respective model capacities. Tiny
reaches approximately 1400 lines per second but is less accurate than the
CRNN. Medium, which achieves the lowest overall error rates, is faster than the
baseline only with batch size 1 and then levels off at approximately 260 lines
per second. On the CPU, only tiny exceeds the CRNN. Even medium nevertheless
remains approximately 45 times faster than Medusa. 

The A40 uses NVIDIA's Ampere architecture; Hopper and Blackwell hardware and
software optimizations are likely to increase PP-OCRv6 throughput relative to
the recurrent baseline.

\section{Conclusion}

PP-OCRv6 can be adapted to historical ATR without changing its basic
architecture or substantially increasing its capacity. The main changes are
an increased input line height, proportional line widths, and the omission of
confidence-window filtering. After these adaptations, from-scratch training on
CATMuS is competitive with the kraken CRNN, while generalized-model pretraining yields
markedly better generalization. The medium model reduces macro CER on the
33-language test set from 12.4 to 7.2\% and also transfers better to
Arabic-script material. Corpus-specific fine-tuning further gives lower error
rates than Medusa on CMMHWR task~1 and CoMMA. Only extreme few-shot adaptation
remains less consistent, with PP-OCRv6 generally requiring 40-80 lines before
improving on the base model.

The adapted models preserve CTC decoding and add no language-generative
mechanism associated with hallucination. The small variant most clearly offers
the free lunch: lower generalized and fine-tuned error rates, fewer parameters,
and comparable or higher GPU throughput than the CRNN. The medium variant
improves accuracy further at the cost of peak batched throughput, but remains
far less expensive than a large VLM.

\section*{Acknowledgements}

This work was funded by the European Union under Grant Agreement No.~101132163
(ATRIUM) and No.~101071829 (MiDRASH). Views and opinions expressed are those of
the authors only and do not necessarily reflect those of the European Union.
This project also received funding from the BPI Scribe project.

\clearpage
{\setlength{\emergencystretch}{1em}%
\printbibliography}

\end{document}